\documentclass[letterpaper, 10 pt, conference]{ieeeconf} %

\IEEEoverridecommandlockouts %

\usepackage[utf8]{inputenc}
\usepackage[T1]{fontenc}
\usepackage{svg}
\usepackage{booktabs}
\usepackage{graphicx}
\usepackage{xcolor}
\usepackage{rotating} 
\usepackage{amssymb} 
\usepackage{amsmath}
\usepackage[table]{xcolor}
\usepackage{array}
\usepackage{capt-of}
\definecolor{barriercolor}{RGB}{255, 120, 50}
\definecolor{bicyclecolor}{RGB}{255, 192, 203}
\definecolor{buscolor}{RGB}{255, 255, 0}
\definecolor{carcolor}{RGB}{0, 150, 245}
\definecolor{constructcolor}{RGB}{0, 255, 255}
\definecolor{motorcolor}{RGB}{200, 180, 0}
\definecolor{pedestriancolor}{RGB}{255, 0, 0}
\definecolor{trafficcolor}{RGB}{255, 240, 150}
\definecolor{trailercolor}{RGB}{135, 60, 0}
\definecolor{truckcolor}{RGB}{160, 32, 240}
\definecolor{driveablecolor}{RGB}{255, 0, 255}
\definecolor{otherflatcolor}{RGB}{139, 137, 137}
\definecolor{sidewalkcolor}{RGB}{75, 0, 75}
\definecolor{terraincolor}{RGB}{150, 240, 80}
\definecolor{manmadecolor}{RGB}{213, 213, 213}
\definecolor{vegetationcolor}{RGB}{0, 175, 0}
\definecolor{otherscolor}{RGB}{0, 0, 0}

\definecolor{rblue}{rgb}{0,0.5,1}
\definecolor{awesome}{rgb}{1.0, 0.13, 0.32}
\definecolor{hollywoodcerise}{rgb}{0.96, 0.0, 0.63}
\definecolor{lasallegreen}{rgb}{0.03, 0.47, 0.19}
\definecolor{hanpurple}{rgb}{0.32, 0.09, 0.98}
\definecolor{green(pigment)}{rgb}{0.0, 0.65, 0.31}
\definecolor{singlegray}{gray}{0.96}
\definecolor{doublegray}{gray}{0.92}
\definecolor{oursgray}{gray}{0.86}

\makeatletter
\let\NAT@parse\undefined
\makeatother
\usepackage[pagebackref=false, breaklinks=true, colorlinks, bookmarks=false]{
        hyperref
}
\hypersetup{
        colorlinks=true,
        linkcolor={red},
        citecolor={hanpurple},
        urlcolor={magenta}
}

\title{\LARGE \bf SelectOccFlow: Selective Spatiotemporal Aggregation for 3D Occupancy and Scene Flow Prediction}

\author{Yuhang Wang$^{1}$, Kai Luo$^{1}$, Yuanfan Zheng$^{1}$, and Kailun Yang$^{1,2,\dag}$%
\thanks{This work was supported in part by the National Natural Science Foundation of China (Grant No. 62473139 and No. 62388101), in part by the Hunan Provincial Research and Development Project (Grant No. 2025QK3019), and in part by the State Key Laboratory of Autonomous Intelligent Unmanned Systems (the opening project number ZZKF2025-2-10).}%
\thanks{$^{1}$The authors are with the School of Artificial Intelligence and Robotics and the National Engineering Research Center of Robot Visual Perception and Control Technology, Hunan University, China (email: kailun.yang@hnu.edu.cn).}%
\thanks{$^{2}$The author is also with the National Engineering Research Center of Robot Visual Perception and Control Technology, Hunan University, China.}%
\thanks{$^{\dag}$Corresponding author: Kailun Yang.}
}
\let\oldtwocolumn\twocolumn
\renewcommand\twocolumn[1][]{%
    \oldtwocolumn[{#1}{
    \begin{center}
    \vskip -3ex
        \centering
        \includegraphics[width=0.99\textwidth]{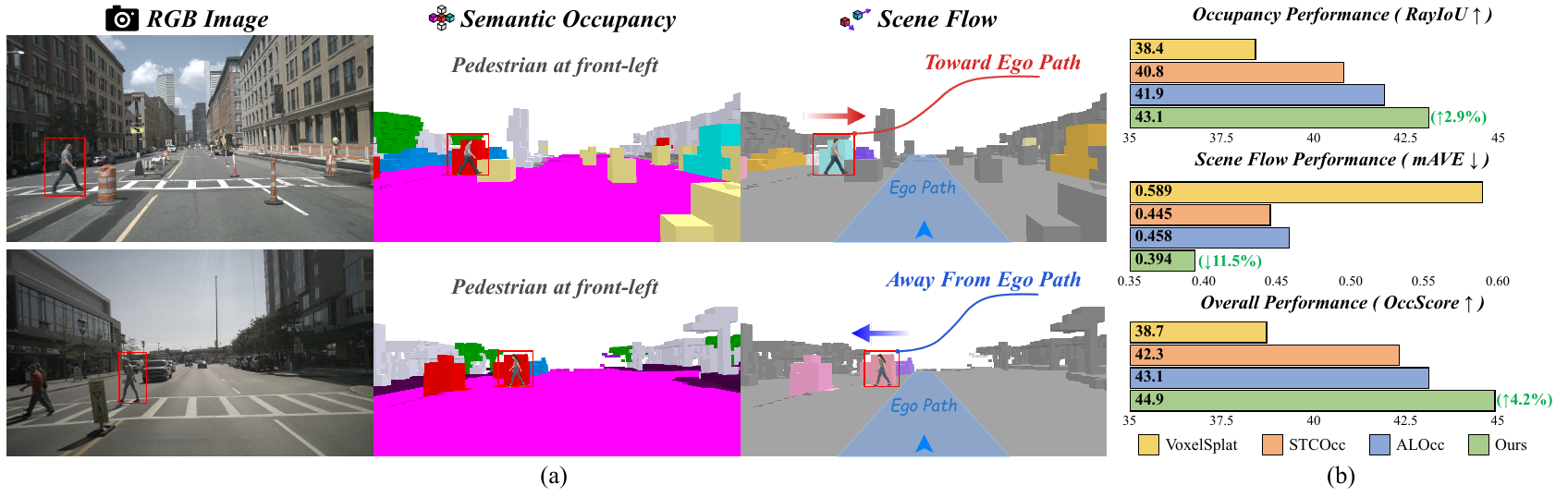}
        \vskip -2ex
        \captionof{figure}{{(a) \textbf{Semantic occupancy and scene flow prediction.}
        Semantic occupancy represents 3D geometry and semantics, while scene flow captures motion. 
        Similar occupancy states can exhibit different scene flows, leading to distinct potential interactions with the ego path, as shown by pedestrians moving toward and away from the path.
        (b) \textbf{Performance comparison on OpenOcc}~\cite{tong2023scene}\textbf{.} SelectOccFlow outperforms VoxelSplat~\cite{zhu2025voxelsplat}, STCOcc~\cite{liao2025stcocc}, and ALOcc~\cite{chen2025alocc} across RayIoU, mAVE, and OccScore, with an OccScore of $44.9$, demonstrating state-of-the-art performance.
        }}
        \label{fig:fig1}
    \end{center}
    }]
}
\begin{document}
        \maketitle
        \thispagestyle{empty}
        \pagestyle{empty}

\begin{abstract}
Comprehensive 3D scene understanding for autonomous driving requires modeling geometry, semantics, and motion. However, camera-based occupancy and scene flow prediction are sensitive to unreliable spatial and temporal aggregation, caused by semantically incompatible image features, misaligned historical observations, and incomplete voxel structures. To address this issue, we propose SelectOccFlow, a selective spatiotemporal aggregation framework that progressively refines contextual evidence across image, temporal, and voxel domains. To obtain semantically compatible image evidence, we design Semantic-Guided Sampling (SGS) to regulate feature sampling with semantic priors. Since reliable image evidence alone cannot resolve temporal inconsistency, we then present State-Conditioned Temporal Aggregation (SCTA) to selectively retrieve historical evidence according to voxel states. To further enhance the structural completeness of voxel representations, we introduce Extent-Aware Spatial Aggregation (ESA), which exploits directional structural support to refine foreground geometry. Experiments on OpenOcc demonstrate that SelectOccFlow achieves a state-of-the-art OccScore of $44.9$, improving the previous best by $+4.2\%$. It also maintains competitive occupancy performance on Occ3D-nus and improves the mean OccScore under nuScenes-C corruptions by $+11.1\%$, demonstrating improved robustness to visual corruptions. The source code will be made publicly available at \url{https://github.com/muchen1021/SelectOccFlow}.
\end{abstract}
        \section{Introduction}
        Reliable autonomous driving requires a comprehensive understanding of the geometry, semantics, and motion of surrounding environments.
Camera-based 3D semantic occupancy prediction provides a dense voxel-level representation of scene geometry and semantic categories~\cite{tong2023scene,behley2019semantickitti,tian2023occ3d}.
However, occupancy primarily describes the instantaneous scene state and does not explicitly characterize motion at occupied voxels.
This limitation is particularly important in dynamic traffic, where geometrically similar scenes can evolve differently.
As illustrated in Fig.~\ref{fig:fig1}(a), similar occupancy states may correspond to different scene flows, implying distinct interactions with the ego path and potential risks.
Scene flow complements occupancy by estimating the voxel-wise direction and magnitude of motion~\cite{liao2025stcocc,chen2025alocc,liu2024let,timoneda2026selfoccflow}.
Together, 3D occupancy and scene flow provide a more complete representation for dynamic scene understanding.

Recovering this representation from multi-view images requires effective aggregation of complementary spatial and temporal evidence.
Existing camera-based approaches extract image features and lift them into BEV or 3D voxel space~\cite{wei2023surroundocc,ma2024cotr,huang2023tri,zhang2023occformer,li2023voxformer}, while temporal extensions incorporate historical observations~\cite{liao2025stcocc,li2024viewformer,ye2024cvt}.
However, contextual evidence is not uniformly reliable.
Spatial evidence can be incompatible across object boundaries, whereas temporal evidence can become unreliable due to independent object motion, changing occlusions, or degraded observations.
Indiscriminate aggregation may therefore propagate errors throughout the prediction pipeline.
The key challenge is thus not only where and how to aggregate spatial and temporal evidence, but also which evidence to select at each stage.

Spatial incompatibility can arise both before and after image-to-3D transformation.
Before lifting, image features near object boundaries or crowded regions may mix foreground, background, and neighboring objects.
Adverse visual conditions can further degrade visual evidence~\cite{kong2023robodepth,xie2025benchmarking}, thereby aggravating feature mixing near object boundaries.
After lifting, incomplete or noisy voxel features require spatial support, while indiscriminate aggregation may propagate errors across semantic boundaries.
Reliable spatial aggregation should therefore preserve semantically compatible image evidence before lifting and exploit structurally consistent voxel support afterward to recover foreground geometry.

Historical observations provide complementary cues for estimating scene dynamics, but their reliability depends on temporal alignment and observation quality.
Although ego-motion compensation aligns static structures, independently moving objects may remain misaligned across frames.
Changing occlusions and degraded visual observations can further produce incomplete or noisy histories~\cite{dong2023benchmarking}.
Moreover, temporal evidence varies across voxel states: stable voxels benefit from longer histories, motion-sensitive voxels require recent observations and adaptive spatial sampling, while free-dominant voxels need only limited temporal support.
A shared temporal range and sampling strategy may retrieve stale, misaligned, or noisy evidence.
Temporal aggregation should therefore adapt historical retrieval to voxel states through state-specific temporal ranges and spatial sampling.

Based on these observations, we propose \textbf{SelectOccFlow}, a selective spatiotemporal aggregation framework for 3D occupancy and scene flow prediction. It adaptively selects context according to spatial compatibility and temporal reliability.
Specifically, Semantic-Guided Sampling (SGS) uses semantic priors to regulate image-space sampling before lifting.
State-Conditioned Temporal Aggregation (SCTA) retrieves historical evidence with state-specific temporal ranges and spatial sampling.
Extent-Aware Spatial Aggregation (ESA) exploits structural voxel support to refine geometry.

Extensive experiments on OpenOcc~\cite{tong2023scene} and Occ3D-nus~\cite{tian2023occ3d} validate the effectiveness of SelectOccFlow for 3D occupancy and scene flow prediction.
As shown in Fig.~\ref{fig:fig1}(b), SelectOccFlow achieves state-of-the-art performance on OpenOcc across occupancy (RayIoU), scene flow (mAVE), and overall evaluation, with an OccScore of $44.9$.
It also maintains competitive occupancy performance on Occ3D-nus.
Under nuScenes-C corruptions~\cite{xie2025benchmarking}, SelectOccFlow further improves the mean OccScore across six corruption types on OpenOcc by $11.1\%$, demonstrating improved robustness under degraded visual conditions.

Our main contributions are summarized as follows:
\begin{itemize}
    \item We propose \textbf{SelectOccFlow}, a selective spatiotemporal aggregation framework that selects evidence based on spatial compatibility and temporal reliability.
    
    \item We develop \textbf{SGS}, \textbf{SCTA}, and \textbf{ESA} for progressive image, temporal, and voxel aggregation, regulating sampling, historical retrieval, and voxel refinement.

    \item Extensive experiments demonstrate state-of-the-art performance on OpenOcc, competitive performance on Occ3D-nus, and improved robustness on nuScenes-C.
\end{itemize}

        \section{Related Work}
        
\subsection{3D Occupancy and Scene Flow Prediction}
Camera-based 3D occupancy prediction methods lift features from monocular images~\cite{li2023voxformer,cao2022monoscene,li2025voxdet} or surround-view observations~\cite{wei2023surroundocc,huang2023tri,zhang2023occformer} into 3D space to estimate voxel-wise occupancy states and semantic categories, providing dense representations of surrounding environments. Recent approaches further extend occupancy prediction through unified panoptic, compact, and sparse formulations~\cite{ma2024cotr,wang2024panoocc,liu2024fully,wang2024opus}. However, occupancy representations primarily describe scene geometry and semantics at a given time without explicitly encoding the motion of occupied voxels.

Scene flow complements occupancy prediction by estimating the motion of occupied voxels over time. Recent methods incorporate scene flow into occupancy frameworks to model semantic geometry and scene dynamics within a unified representation~\cite{zhu2025voxelsplat,chen2025alocc,liu2024let,timoneda2026selfoccflow,li2024viewformer}.
Despite this progress, occupancy and scene flow remain closely coupled because inaccurate spatial structures can impair motion estimation, while unreliable temporal cues may hinder the recovery of both 3D occupancy and scene flow. Consequently, reliable dynamic scene understanding requires effective modeling of spatial observations and temporal evidence.

\subsection{Spatiotemporal Modeling}
Spatiotemporal modeling has been widely explored in camera-based 3D perception to enhance scene representation and dynamic understanding~\cite{liao2025stcocc,li2024viewformer,li2024bevformer,qin2023unifusion}. 
Existing methods model spatial context through dense BEV or voxel representations~\cite{wei2023surroundocc,zhang2023occformer,li2024bevformer} and sparse query-based aggregation~\cite{wang2024opus,wang2022detr3d,tang2024sparseocc}.
Dense representations aggregate context over structured grids, whereas sparse queries retrieve features around task-specific reference locations. However, spatial compatibility during feature aggregation before and after lifting remains less explored, and evidence from a single frame may be incomplete or ambiguous.

\begin{figure*}[!t]
    \centering
    \includegraphics[width=\textwidth]{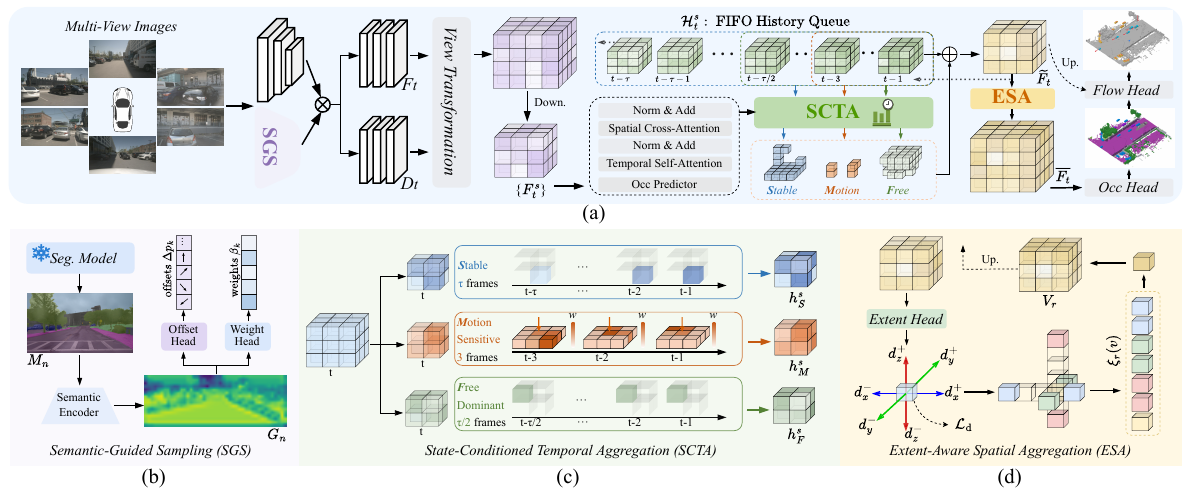}
    \vskip-2ex
    \caption{
    \textbf{Overview of SelectOccFlow.} 
    (a) Overall pipeline for 3D occupancy and scene flow prediction. 
    (b) SGS performs semantic-guided image-space sampling before lifting.
    (c) SCTA adapts historical aggregation to voxel states.
    (d) ESA exploits directional extents for spatial feature refinement.
    }
    \vskip-3ex
    \label{fig:fig2}
\end{figure*}

Temporal modeling mitigates this limitation by incorporating historical observations to complement current evidence and capture scene evolution.
Existing methods encode temporal context through recurrent BEV features~\cite{li2024bevformer,chang2024recurrentbev}, aligned multi-frame perspective features~\cite{qin2023unifusion,liu2023petrv2}, or propagated object queries~\cite{wang2023streampetr,papais2025foresight}. For occupancy prediction, recent approaches further employ multi-frame streaming temporal attention~\cite{li2024viewformer}, perform sparse voxel updates~\cite{liao2025stcocc}, or introduce historical predictions as memory priors~\cite{leng2025occupancy,yuan2025lmpocc}.
Although effective, these methods determine how history is represented and fused, while variations in the relevance of historical evidence across voxel states remain insufficiently addressed.
Degraded visual observations caused by adverse weather, illumination changes, and sensor failures~\cite{kong2023robodepth,xie2025benchmarking,dong2023benchmarking,wu2025see,jiang2025acdepth} can weaken spatial evidence and reduce the reliability of historical cues, while scene dynamics increase discrepancies between current and historical evidence.
        \section{Methodology}
        
\subsection{Overall Architecture}
\label{sec:overview}

As illustrated in Fig.~\ref{fig:fig2}, SelectOccFlow follows a lift-and-refine pipeline for 3D occupancy and scene flow prediction from multi-view images. Given $N_I$ input images $I \in \mathbb{R}^{N_I \times H \times W \times 3}$, the model is supervised by an occupancy grid $O_c \in \mathbb{R}^{X \times Y \times Z}$ and a scene flow grid $O_f \in \mathbb{R}^{X \times Y \times Z \times 2}$, encoding voxel semantics and planar motion, respectively.

Semantic-Guided Sampling (SGS) first refines multi-view image features before lifting them into voxel space (see Sec.~\ref{sec:1}).
The lifted features are encoded by the OA Transformer~\cite{liao2025stcocc} into multi-scale voxel representations $\{F_t^s\}$, where $s\in\{1/8,1/4,1/2\}$.
State-Conditioned Temporal Aggregation (SCTA) incorporates historical evidence at each scale and propagates the enhanced features $\widetilde{F}_t^s$ from coarse to fine (see Sec.~\ref{sec:2}).
The finest-scale feature $\widetilde{F}_t$ is further refined by Extent-Aware Spatial Aggregation (ESA) for occupancy prediction (see Sec.~\ref{sec:3}), producing $\overline{F}_t$.
Occupancy cues and $\widetilde{F}_t$ are fused for scene flow prediction.

\subsection{Semantic-Guided Sampling}
\label{sec:1}

Standard image backbones aggregate local evidence using fixed sampling patterns without explicitly considering scene semantics. Near semantic boundaries, a receptive field may cover incompatible regions and mix responses from different categories. Such ambiguity can propagate into voxel space through view transformation. As illustrated in Fig.~\ref{fig:fig2}(b), we introduce Semantic-Guided Sampling (SGS), which uses categorical semantic priors to adaptively regulate image-space sampling
locations and response weights before lifting.

For the $n$-th camera view, let $F_n^I \in \mathbb{R}^{C \times H' \times W'}$ denote the backbone feature. We use a frozen YOLO26 semantic segmentation model~\cite{jocher2026ultralytics} to obtain a semantic map $M_n$. To preserve spatial alignment, $M_n$ undergoes the same geometric transformations as the corresponding RGB image. The category labels are then embedded and processed by a lightweight semantic encoder to obtain a guidance feature $G_n$, which is spatially aligned with $F_n^I$ and provides semantic guidance for predicting adaptive sampling parameters.

Given $G_n$, a lightweight sampling predictor $\mathcal{P}$ outputs $K=9$ offsets $\Delta\mathbf{p}_k(\mathbf{p})$ and corresponding modulation weights $\beta_k(\mathbf{p})$ at each spatial location $\mathbf{p}$.
Each offset is bounded within $\pm2$ cells along both spatial dimensions of the $H'\times W'$ feature map, while the modulation weights control the contribution of individual sampled responses.
For the $k$-th sampling point, the adaptive location is $\mathbf{q}_k(\mathbf{p}) = \mathbf{p}+\mathbf{p}_k+\Delta\mathbf{p}_k(\mathbf{p})$, where $\mathbf{p}_k$ denotes the regular offset of the $k$-th point in a $3\times3$ sampling grid centered at $\mathbf{p}$.

SGS aggregates responses sampled from $F_n^I$ at the adaptive locations using the predicted modulation weights:
\begin{equation}
    \widetilde{F}_n^I(\mathbf{p})
    =
    F_n^I(\mathbf{p})
    +
    \sum_{k=1}^{K}
    \beta_k(\mathbf{p}) W_k
    \mathcal{S}_{\mathrm{2D}}
    \left(
        F_n^I,\mathbf{q}_k(\mathbf{p})
    \right),
    \label{eq:sgs}
\end{equation}
where $\mathcal{S}_{\mathrm{2D}}(\cdot)$ denotes bilinear sampling and $W_k$ denotes the learnable kernel weight of the $k$-th sampling point.
The residual connection preserves the original feature while incorporating the adaptively sampled context.

The frozen segmentation model, pretrained on the 19-class Cityscapes dataset~\cite{cordts2016cityscapes} without target-domain fine-tuning, only guides sampling offsets and weights. All sampled values come from the RGB feature $F_n^I$, with no semantic-model features fused into the occupancy representation.
The refined feature $\widetilde{F}_n^I$ is then lifted into voxel space through view transformation. The resulting voxel feature is downsampled and encoded by the OA Transformer at each scale, producing the multi-scale representations $\{F_t^s\}$ for temporal aggregation.

\subsection{State-Conditioned Temporal Aggregation}
\label{sec:2}
Temporal evidence varies in persistence and correspondence reliability across voxel states. Stable regions benefit from long-term aligned history, whereas motion-sensitive regions require local correspondence search and free-dominant regions need only lightweight temporal support. We therefore introduce State-Conditioned Temporal Aggregation (SCTA) to adapt historical aggregation to the current voxel state.

Given the multi-scale voxel representations $\{F_t^s\}$, SCTA operates at each scale with a FIFO history queue $\mathcal{H}_t^s$ that stores detached fused voxel features $H_{t-\ell}^s$ and occupancy cues in their original coordinate frames. During retrieval, selected current voxels are transformed to the corresponding historical frames through recursively composed ego-motion transforms and sampled from $H_{t-\ell}^s$ by trilinear interpolation.

Let $\boldsymbol{\pi}_t^s(\mathbf{v})$ denote the semantic class-probability vector at voxel $\mathbf{v}$. The non-empty probability, movable-class probability, and prediction confidence are $o_t^s(\mathbf{v})=1-\pi_{t,c_{\mathrm f}}^s(\mathbf{v})$, $m_t^s(\mathbf{v})=\sum_{c\in\mathcal{C}_{\mathrm m}} \pi_{t,c}^s(\mathbf{v})$, and $p_t^s(\mathbf{v})=\max_c\pi_{t,c}^s(\mathbf{v})$, where $c_{\mathrm f}$ denotes the free-space class and $\mathcal{C}_{\mathrm m}$ the set of movable classes.
At each scale, top-ranked voxels by $o_t^s$ are selected as occupied candidates and routed by the following score:
\begin{equation}
    q_t^s(\mathbf{v})
    =
    m_t^s(\mathbf{v})o_t^s(\mathbf{v})
    \left(
    1+
    \left|
    o_t^s(\mathbf{v})
    -
    \widetilde{o}_{t-1}^s(\mathbf{v})
    \right|
    \right),
    \label{eq:scta_route}
\end{equation}
where $\widetilde{o}_{t-1}^s(\mathbf{v})$ denotes the previous-frame non-empty probability aligned to the current frame. The score combines occupancy likelihood, class movability, and temporal variation.
Under a fixed quota, occupied candidates with the highest $q_t^s$ form the motion-sensitive set $\Omega_{\mathrm M}^s$, while the remaining occupied candidates form the stable set $\Omega_{\mathrm S}^s$; voxels outside the occupied candidates constitute the free-dominant set $\Omega_{\mathrm F}^s$.

The three states use distinct temporal windows and retrieval strategies (Fig.~\ref{fig:fig2}(c)). For a motion-sensitive voxel $\mathbf{v}$, let $\mathbf{x}_{\ell}^{s}(\mathbf{v})$ denote its ego-motion-compensated location in frame $t-\ell$. To account for residual object motion, candidate features are sampled using a scale-dependent offset set $\mathcal{O}_{s}$:
\begin{equation}
    \mathbf{h}_{\ell,j}^{s}(\mathbf{v})
    =
    \mathcal{S}_{\mathrm{3D}}
    \left(
        H_{t-\ell}^{s},
        \mathbf{x}_{\ell}^{s}(\mathbf{v})
        +
        \boldsymbol{\delta}_{j}^{s}
    \right),
    \qquad
    \boldsymbol{\delta}_{j}^{s}\in\mathcal{O}_{s},
    \label{eq:scta_surface}
\end{equation}
where $\mathcal{S}_{\mathrm{3D}}(\cdot)$ denotes trilinear sampling.
The candidates are adaptively aggregated within each frame and concatenated
over the motion-sensitive temporal window $\mathcal{T}_{\mathrm M}^{s}$ as follows:
\begin{equation}
    \mathbf{h}_{\mathrm{M}}^{s}(\mathbf{v})
    =
    \left[
    \sum_j
    \operatorname{softmax}_{j}
    \left(
        \phi(\mathbf{c}_{\ell,j}^{s}(\mathbf{v}))
    \right)
    \mathbf{h}_{\ell,j}^{s}(\mathbf{v})
    \right]_{\ell\in\mathcal{T}_{\mathrm{M}}^{s}}.
    \label{eq:scta_surface_attn}
\end{equation}
The descriptor $\mathbf{c}_{\ell,j}^{s}(\mathbf{v})$ encodes current--historical feature similarity, historical non-empty and movable-class probabilities, offset distance, and validity. $\phi(\cdot)$ maps it to a scalar score, with invalid candidates masked before softmax.

Stable occupied voxels aggregate the full ego-aligned history using frame-wise reliability weights derived from the current non-empty probability, confidence, feature discrepancy, sampling validity, and temporal distance. Free-dominant voxels instead use ego-aligned point sampling over a shorter window without frame-wise weighting. Let $\mathbf{h}_{a}^{s}$ denote the resulting history for $a\in\{\mathrm{S},\mathrm{M},\mathrm{F}\}$, and $F_{t,a}^{s}$ and $\mathbf{z}_{t,a}^{s}$ the current voxel features and occupancy embeddings at $\Omega_a^s$. The temporal representation is encoded as follows:
\begin{equation}
    \mathbf{u}_{a}^{s}
    =
    \Phi_{a}^{s}
    \left(
        [
        F_{t,a}^{s},
        \mathbf{h}_{a}^{s},
        \mathbf{z}_{t,a}^{s}
        ]
    \right),
    \label{eq:scta_encode}
\end{equation}
where $\Phi_a^s$ captures the temporal characteristics of state $a$, and a shared projection $\Psi^s$ maps the encoded representations into a common feature space. An update gate $\mathbf{g}_a^s$, derived from the non-empty probability, temporal occupancy discrepancy, and confidence, modulates the state-specific temporal residual:
\begin{equation}
    \widetilde{F}_{t}^{s}
    =
    F_{t}^{s}
    +
    \sum_{a\in\{\mathrm{S},\mathrm{M},\mathrm{F}\}}
    \operatorname{Scatter}_{\Omega_{a}^{s}}
    \left(
        \mathbf{g}_{a}^{s}
        \odot
        \Psi^{s}(\mathbf{u}_{a}^{s})
    \right),
    \label{eq:scta_update}
\end{equation}
where $\operatorname{Scatter}_{\Omega_a^s}(\cdot)$ writes the residuals to their corresponding voxel locations. The detached fused feature $\widetilde{F}_t^s$ and occupancy cues are then inserted into $\mathcal{H}_t^s$, replacing the oldest entry when the queue is full. At coarser scales, $\widetilde{F}_t^s$ is progressively upsampled for coarse-to-fine refinement. The finest-scale output $\widetilde{F}_t$ is then used for spatial refinement.

\begin{table*}[!t]
    \centering
    \caption{
    Comparison of RayIoU (\%), mAVE, and OccScore on OpenOcc~\cite{tong2023scene}. $^{\dagger}$ denotes results reproduced with the official code and evaluated under the same protocol. Sup.: C/L/3D denote camera, LiDAR, and 3D occupancy supervision.}
    \vskip-1ex
    \resizebox{\textwidth}{!}{
    \begin{tabular}{l| c c c |c c c |>{\columncolor{gray!10}}c >{\columncolor{gray!10}}c >{\columncolor{gray!10}}c}
        \toprule Method  &Sup.  & Backbone & Input Size     & RayIoU$_{1m}$ (\%) $\uparrow$ & RayIoU$_{2m}$ (\%) $\uparrow$ & RayIoU$_{4m}$ (\%) $\uparrow$ & RayIoU (\%) $\uparrow$ & mAVE $\downarrow$ & OccScore $\uparrow$ \\
        \midrule 
        OccNeRF-L~\cite{zhang2025occnerf}    & C+L & ResNet-101      & 256$\times$704   & 16.6 & 29.3 & 49.2 & 31.7 & 1.59  & 28.5\\
        RenderOcc~\cite{pan2024renderocc}    & L   & Swin-Base       & 512$\times$1408  & 20.3 & 32.7 & \underline{49.9} & 36.7 & 1.63  & 33.0\\
        LetOccFlow-L~\cite{liu2024let}       & C+L & ConvNeXt-Base   & 512$\times$1408  & 25.5 & 39.7 & \textbf{56.3} & 40.5 & 1.45  & 36.4\\
        CascadeFlow~\cite{liao2024cascadeflow}  & 3D  & ResNet-50    & 256$\times$704   & 33.5 & 40.6 & 45.3 & 39.6 & 0.47  & 40.9\\
        F-Occ~\cite{zhao20243d}              & 3D  & Intern-T        & 256$\times$704   & 33.9 & 40.7 & 45.2 & 39.9 & 0.493 & 41.0\\
        FB-Occ$^{\dagger}$~\cite{li2023fb}   & 3D  & ResNet-50       & 256$\times$704   & 31.9 & 39.2 & 43.7 & 38.3 & 0.954 & 34.9\\
        VoxelSplat$^{\dagger}$~\cite{zhu2025voxelsplat}  & 3D  & ResNet-50       & 256$\times$704   & 31.8 & 39.3 & 44.1 & 38.4 & 0.589 & 38.7\\
        STCOcc~\cite{liao2025stcocc}         & 3D  & ResNet-50       & 256$\times$704   & 35.0 & 41.7 & 45.9 & 40.8 & 0.445 & 42.3 \\
        ALOcc-2D~\cite{chen2025alocc}        & 3D  & ResNet-50       & 256$\times$704   & 34.3 & 41.0 & 45.5 & 40.3 & \underline{0.431} & 41.9 \\
        ALOcc~\cite{chen2025alocc}        & 3D  & ResNet-50       & 256$\times$704   & \underline{35.6} & \underline{42.9} & 47.2 & \underline{41.9} & 0.458 & \underline{43.1} \\
        \midrule                          
        Ours&3D & ResNet-50 & 256$\times$704 & \textbf{37.3} & \textbf{44.0} & 48.0 & \textbf{43.1} & \textbf{0.394} & \textbf{44.9} \\
        \bottomrule
    \end{tabular}}
    \label{table:OpenOcc_main_results}
    \vskip-3ex
\end{table*}

\begin{table}
\centering
\caption{Comparison of RayIoU (\%) performance on Occ3D-nus~\cite{tian2023occ3d}. All methods are trained without the camera-visible mask.}
\vskip-1ex
\resizebox{\columnwidth}{!}
{
\begin{tabular}{l|c c|c c c|>{\columncolor{gray!10}}c}
\toprule
Method & Backbone & Input Size 
& \multicolumn{3}{|c|}{RayIoU$_{1,2,4m}$ $\uparrow$} 
& RayIoU $\uparrow$ \\
\midrule
BEVFormer~\cite{li2024bevformer}        & R101  &900$\times$1600    & 26.1 & 32.9 & 38.0 & 32.4 \\
RenderOcc~\cite{pan2024renderocc}       &Swin-B &512$\times$1408    & 13.4 & 19.6 & 25.5 & 19.5\\
BEVDet-Occ~\cite{huang2022bevdet4d}     & R50   & 256$\times$704    & 23.6 & 30.0 & 35.1 & 29.6\\
FB-Occ~\cite{li2023fb}                  & R50   & 256$\times$704    & 33.0 & 39.9 & 44.0 & 39.0\\
SparseOcc~\cite{tang2024sparseocc} & R50   & 256$\times$704    & 29.1 & 35.8 & 40.3 & 35.1\\
COTR~\cite{ma2024cotr}                  & R50   & 256$\times$704    & 36.3 & 41.7 & 45.1 & 41.0\\
OPUS-L~\cite{wang2024opus}              & R50   & 256$\times$704    & 34.7 & 42.1 & 46.7 & 41.2\\
STCOcc~\cite{liao2025stcocc}            & R50   & 256$\times$704    & 36.2 & 42.7 & 46.4 & 41.7 \\
STCOcc~\cite{liao2025stcocc}            & R50   & 512$\times$1408   & 36.9 & 42.8 & 46.7 & 42.1 \\
ALOcc-2D~\cite{chen2025alocc}           & R50   & 256$\times$704    & 37.1 & 43.8 & \underline{48.2} & 43.0  \\
ALOcc~\cite{chen2025alocc}           & R50   & 256$\times$704    & \textbf{37.8} & \textbf{44.7}  & \textbf{48.8} & \textbf{43.7} \\
\midrule
Ours                                    & R50   & 256$\times$704    & \textbf{37.8} & \underline{44.2} & 47.9 & \underline{43.3} \\
\bottomrule
\end{tabular}}
\label{table:Occ3D_rayiou}
\vskip-3ex
\end{table}
\subsection{Extent-Aware Spatial Aggregation}
\label{sec:3}
After temporal aggregation, $\widetilde{F}_t$ may still lack fine-grained
structural support for movable objects. We therefore introduce Extent-Aware
Spatial Aggregation (ESA) to exploit directional extents for spatial refinement (see Fig.~\ref{fig:fig2}(d)).

Inspired by the six-direction boundary representation in
VoxDet~\cite{li2025voxdet}, we characterize each occupied voxel along
$\mathcal{D}
=
\{
\mathrm{x}^{+},
\mathrm{x}^{-},
\mathrm{y}^{+},
\mathrm{y}^{-},
\mathrm{z}^{+},
\mathrm{z}^{-}
\}$.
Let $c(\mathbf{v})$ denote the category label of voxel $\mathbf{v}=(x,y,z)$.
For each direction $r\in\mathcal{D}$, the directional extent is defined as follows:
\begin{equation}
    d_r(\mathbf{v})
    =
    \min
    \left\{
        k>0 :
        c(\mathbf{v}+k\mathbf{e}_r)
        \neq c(\mathbf{v})
    \right\},
    \label{eq:esa_extent}
\end{equation}
where $\mathbf{e}_r$ denotes the unit vector along the direction $r$. The search is restricted to the valid voxel grid and truncated at the boundary if no category change occurs. During training, $d_r(\mathbf{v})$ is normalized by the grid size along the corresponding axis. The normalized prediction is rescaled to $\widehat{d}_r(\mathbf{v})$, defining the support location as
$\boldsymbol{\xi}_r(\mathbf{v})
=
\mathbf{v}
+
\widehat{d}_r(\mathbf{v})\mathbf{e}_r$.

ESA trilinearly samples six support features and aggregates them according to their relevance to the current voxel. Let $Q(\mathbf{v})$ denote the query projection of $\widetilde{F}_t(\mathbf{v})$, and $K_r(\mathbf{v})$ and $V_r(\mathbf{v})$ the key and value projections of the $r$-th support feature. The directional attention weight is defined as
\begin{equation}
    \alpha_r(\mathbf{v})
    =
    \operatorname{softmax}_{r}
    \left(
        \frac{
            Q(\mathbf{v})^{\top}K_r(\mathbf{v})
        }{
            \sqrt{C_v}
        }
    \right),
    \label{eq:esa_weight}
\end{equation}
where $C_v$ denotes the voxel feature dimension. The final spatially refined voxel feature is computed as
\begin{equation}
    \overline{F}_{t}(\mathbf{v})
    =
    \Gamma
    \left(
        \widetilde{F}_{t}(\mathbf{v})
        +
        \sum_{r\in\mathcal{D}}
        \alpha_r(\mathbf{v})V_r(\mathbf{v})
    \right),
    \label{eq:esa_aggregation}
\end{equation}
where $\Gamma(\cdot)$ denotes the output projection.

Movable classes generally correspond to discrete foreground structures with clear spatial extents, whereas static classes such as roads and vegetation often form continuous regions without explicit boundaries. Since directional extents provide more informative structural cues for foreground structures, we retain supervision for all occupied voxels while assigning greater weight to movable voxels and excluding empty voxels with no valid extent. Accordingly, we compute the mean extent errors over movable and static occupied voxels separately and define the extent loss as
\begin{equation}
    \mathcal{L}_{\mathrm d}
    =
    \frac{
        \rho\mathcal{L}_{\mathrm d}^{(1)}
        +
        \mathcal{L}_{\mathrm d}^{(2)}
    }{
        \rho+1
    },
    \label{eq:esa_loss}
\end{equation}
where $\mathcal{L}_{\mathrm d}^{(1)}$ denotes the mean $\ell_1$ extent error over movable occupied voxels, $\mathcal{L}_{\mathrm d}^{(2)}$ denotes that over static occupied voxels, and $\rho$ controls their relative weighting. This formulation emphasizes foreground geometry while retaining structural supervision for static regions to refine voxel spatial structure.

\subsection{Loss Functions}
\label{sec:loss}
The representations at the finest scale support the two prediction branches. Occupancy logits from $\overline{F}_t$ are upsampled to the full voxel resolution, while $\widetilde{F}_t$ is first upsampled and combined with occupancy cues for scene flow prediction.

Training is supervised by depth, occupancy, directional extent, and scene flow objectives. Auxiliary occupancy losses are applied to the multi-scale voxel representations, while $\mathcal{L}_{\mathrm{occ}}$ supervises the final occupancy prediction. The complete training objective is formulated as follows:
\begin{equation}
    \mathcal{L}
    =
    \mathcal{L}_{\mathrm{depth}}
    +
    \sum_s
    \gamma_s
    \mathcal{L}_{\mathrm{occ}}^{s}
    +
    \mathcal{L}_{\mathrm{occ}}
    +
    \mathcal{L}_{\mathrm d}
    +
    \mathcal{L}_{\mathrm{flow}},
    \label{eq:total_loss}
\end{equation}
where $\gamma_s$ controls the auxiliary occupancy supervision at scale $s$. All other loss terms use unit weights.

        \section{Experiments}
        \subsection{Experimental Setup}
\noindent\textbf{Datasets.}
We primarily evaluate SelectOccFlow on OpenOcc~\cite{tong2023scene}, which provides annotations for both 3D semantic occupancy and scene flow and supports joint training and evaluation of the two tasks. We further report occupancy results on Occ3D-nus~\cite{tian2023occ3d} to complement the evaluation of 3D semantic occupancy. Since scene flow annotations are unavailable, the flow branch is removed on this benchmark. Both benchmarks are derived from nuScenes~\cite{caesar2020nuscenes} and follow its standard training and validation splits. To further assess robustness to visual corruptions, we evaluate the model on nuScenes-C~\cite{xie2025benchmarking}, which applies six synthetic corruption types at three severity levels to the nuScenes validation set.

\noindent\textbf{Metrics.}
We follow the standard evaluation protocol for each benchmark. For 3D semantic occupancy, we report Ray-based Intersection over Union (RayIoU)~\cite{liu2024fully} and mean Intersection over Union (mIoU). RayIoU is evaluated at distance thresholds of $1$, $2$, and $4$ meters, and the three results are averaged for the final score. Scene flow on OpenOcc~\cite{tong2023scene} is evaluated using mean absolute velocity error (mAVE) over true positive occupied voxels of the evaluated movable classes (\textit{e.g.}, \emph{car} and \emph{pedestrian}). 
Following the CVPR $2024$ Autonomous Grand Challenge protocol, OccScore~\cite{zhao20243d} jointly evaluates 3D semantic occupancy and scene flow:
\begin{equation}
    \mathrm{OccScore}
    =
    0.9\,\mathrm{RayIoU}
    +
    0.1\,\max(1-\mathrm{mAVE},0).
\end{equation}

\begin{table*}
    \centering
    \caption{Comparison of mIoU (\%) performance on Occ3D-nus~\cite{tian2023occ3d}. $\dagger$ denotes results obtained by retraining with the official code. $^\ast$~indicates models trained without the camera-visible mask.}
    \vskip-1ex
    \resizebox{\textwidth}{!}{
    \begin{tabular}{l|cc|>{\columncolor{gray!10}}c |ccccccccccccccccc}
        \toprule \rotatebox{90}{Method}     & \rotatebox{90}{Backbone}                                               %
                                            & \rotatebox{90}{Input Size}                                             %
                                            & \rotatebox{90}{mIoU (\%)}                                               %
                                            & \rotatebox{90}{\textcolor{otherscolor}{$\blacksquare$} others}         %
                                            & \rotatebox{90}{\textcolor{barriercolor}{$\blacksquare$} barrier}       %
                                            & \rotatebox{90}{\textcolor{bicyclecolor}{$\blacksquare$} bicycle}       %
                                            & \rotatebox{90}{\textcolor{buscolor}{$\blacksquare$} bus}               %
                                            & \rotatebox{90}{\textcolor{carcolor}{$\blacksquare$} car}               %
                                            & \rotatebox{90}{\textcolor{constructcolor}{$\blacksquare$} cons. veh.}  %
                                            & \rotatebox{90}{\textcolor{motorcolor}{$\blacksquare$} motor.}          %
                                            & \rotatebox{90}{\textcolor{pedestriancolor}{$\blacksquare$} pedes.}     %
                                            & \rotatebox{90}{\textcolor{trafficcolor}{$\blacksquare$} tfc. cone}     %
                                            & \rotatebox{90}{\textcolor{trailercolor}{$\blacksquare$} trailer}       %
                                            & \rotatebox{90}{\textcolor{truckcolor}{$\blacksquare$} truck}           %
                                            & \rotatebox{90}{\textcolor{driveablecolor}{$\blacksquare$} drv. surf.}  %
                                            & \rotatebox{90}{\textcolor{otherflatcolor}{$\blacksquare$} other flat}  %
                                            & \rotatebox{90}{\textcolor{sidewalkcolor}{$\blacksquare$} sidewalk}     %
                                            & \rotatebox{90}{\textcolor{terraincolor}{$\blacksquare$} terrain}       %
                                            & \rotatebox{90}{\textcolor{manmadecolor}{$\blacksquare$} manmade}       %
                                            & \rotatebox{90}{\textcolor{vegetationcolor}{$\blacksquare$} vegetation} \\
        \midrule
        BEVFormer$^\ast$~\cite{li2024bevformer}&R101& 900$\times$1600 & 26.9 & 5.9 & 37.8 & 17.9 & 40.4 & 42.4 & 7.4 & 23.9 & 21.8 & 21.0 & 22.4 & 30.7 & 55.4 & 28.4 & 36.0 & 28.1 & 20.0 & 17.7 \\
        CTF-Occ$^\ast$~\cite{tian2023occ3d}&R101& 900$\times$1600 &28.5 & 8.1 & 39.3 & 20.6 & 38.3 & 42.2 & 16.9 & 24.5 & 22.7 & 21.1 & 23.0 & 31.1 & 53.3 & 33.8 & 38.0 & 33.2 & 20.8 & 18.0 \\
        TPVFormer$^\ast$~\cite{huang2023tri}&R101& 900$\times$1600 &27.8 & 7.2 & 38.9 & 13.7 & 40.8 & 45.9 & 17.2 & 20.0 & 18.9 & 14.3 & 26.7 & 34.2 & 55.7 & 35.5 & 37.6 & 30.7 & 19.4 & 16.8 \\
        OSP~\cite{shi2024occupancy}  & R101 & 900$\times$1600  & 41.2 & 11.0 & 49.0 & 27.7 & 50.2 & 56.0 & 23.0 & 31.0 & 30.9 & 30.3 & 35.6 & 41.2 & 82.1 & 42.6 & 51.9 & 55.1 & 44.8 & 38.2 \\
        FB-Occ~\cite{li2023fb} & R50 & 256$\times$704 & 39.8 & 13.8 & 44.5 & 27.1 & 46.2 & 49.7 & 24.6 & 27.4 & 28.5 & 28.2 & 33.7 & 36.5 & 81.7 & 44.1 & 52.6 & 56.9 & 42.6 & 38.1 \\
        ViewFormer~\cite{li2024viewformer}& R50  & 256$\times$704  & 41.9 & 12.9 & 50.1 & 27.9 & 44.6 & 52.9 & 22.4 & 29.6 & 28.0 & 29.2 & 35.2 & 39.4 & 84.7 & 49.4 & 57.4 & 59.7 & 47.4 & 40.6 \\
        COTR$^{\dagger}$~\cite{ma2024cotr} & R50 & 256$\times$704  & 40.4 & 11.6 & 47.4 & 28.0 & 42.3 & 52.3 & 26.4 & 28.9 & 28.1 & 31.3 & 33.9 & 39.0 & 81.6 & 42.0 & 52.5 & 54.6 & 46.7 & 40.9 \\
        STCOcc$^{\dagger}$~\cite{liao2025stcocc} & R50 & 256$\times$704 & 42.2 & 14.0 & 48.0 & 26.8 & 45.7 & 52.3 & 29.0 & 30.1 & 30.1 & 30.9 & 36.8 & 40.8 & 82.2 & 45.6 & 54.8 & 59.0 & 49.2 & 42.1 \\
        ALOcc~\cite{chen2025alocc} & R50 & 256$\times$704  & 45.5 & 15.3 & 52.5 & 30.8 & 47.2 & 55.9 & 32.7 & 33.3 & 32.4 & 36.2 & 38.9 & 43.7 & 84.9 & 48.5 & 58.8 & 61.9 & 53.5 & 47.3 \\
        \midrule 
        Ours & R50  & 256$\times$704  & 44.6 & 15.1 & 51.1 & 30.4 & 47.1 & 55.4 & 30.3 & 31.7 & 32.4 & 34.7 & 39.8 & 43.4 & 84.7 & 47.5 & 57.5 & 61.1 & 51.3 & 44.6 \\        
        \bottomrule
    \end{tabular}
    } \label{table:Occ3D_miou}
    \vskip-3ex
\end{table*}

\noindent\textbf{Implementation Details.}
We use $256\times704$ images with a ResNet-50~\cite{he2016deep} backbone and BEVStereo~\cite{li2023bevstereo} for depth estimation. Following prior works~\cite{liao2025stcocc,chen2025alocc,li2023bevstereo,li2023bevdepth}, sparse LiDAR points provide depth supervision only during training. The 3D space is discretized into a $200\times200\times16$ voxel grid with $0.4\,\mathrm{m}$ resolution.
For SCTA, stable, motion-sensitive, and free-dominant voxels use $\tau_s$, $3$, and $\tau_s/2$ frames, respectively, with $\tau_s=4,8,16$ and $5,9,13$ offsets from coarse to fine.
For ESA, $\rho=5$. The auxiliary occupancy weights are $\gamma_{1/8}=1/8$, $\gamma_{1/4}=1/4$, and $\gamma_{1/2}=1/2$, while all other losses use unit weights. Unless otherwise specified, our models are trained with AdamW~\cite{loshchilov2017decoupled} using an initial learning rate of $5\times10^{-5}$, a weight decay of $0.01$, and a cosine learning rate schedule. Training uses a batch size of $8$ on four NVIDIA RTX 3090 GPUs, and the training schedule for each benchmark follows its corresponding baseline setting.

\subsection{Main Results}
We evaluate our method on OpenOcc~\cite{tong2023scene} for 3D occupancy and scene flow prediction, and on Occ3D-nus~\cite{tian2023occ3d} for 3D occupancy prediction.
We further assess robustness under diverse visual corruptions using nuScenes-C~\cite{xie2025benchmarking}.
We use officially released checkpoints when available and retrain the remaining methods following their reported configurations.

\noindent\textbf{Main Results on OpenOcc.}
As shown in Tab.~\ref{table:OpenOcc_main_results}, our method achieves the best overall performance on the OpenOcc validation set~\cite{tong2023scene}, with an OccScore of $44.9$, surpassing the previous best by $1.8$.
Compared with the best previously reported result for each metric, it improves RayIoU from $41.9$ to $43.1$ and reduces mAVE from $0.431$ to $0.394$, achieving the best results for both 3D occupancy and scene flow prediction.
The simultaneous gains indicate that our method more accurately captures both spatial occupancy structure and voxel-wise motion, providing richer geometry and motion cues for understanding dynamic traffic participants.

\begin{table}[t]
\centering
\caption{Robustness comparison on OpenOcc~\cite{tong2023scene} and Occ3D-nus~\cite{tian2023occ3d} under nuScenes-C~\cite{xie2025benchmarking} degradations. $\dagger$ denotes results obtained by retraining with the official code.}
\label{tab:nuscenes_c_robustness}
\vskip-1ex
\resizebox{\columnwidth}{!}{
\begin{tabular}{l|cccccc|>{\columncolor{gray!10}}c}
\toprule
Method & Bright & Dark & Fog & Snow & Motion & Quant & Avg. \\
\midrule
\multicolumn{8}{c}{\textit{OpenOcc}~\cite{tong2023scene}~\textit{OccScore $\uparrow$}}\\
\midrule
FB-Occ$^{\dagger}$~\cite{li2023fb}    & 28.0 & 17.0 & 27.3 & 5.3  & 24.4 & 22.8 & 20.8 \\
VoxelSplat$^{\dagger}$~\cite{zhu2025voxelsplat} & 29.5 & 17.4 & 29.9 & 3.7 & 22.8 & 22.6 & 21.0 \\
STCOcc~\cite{liao2025stcocc} & 33.3 & 18.1 & 32.7 & 9.8  & 29.5 & 27.6 & 25.2 \\
ALOcc~\cite{chen2025alocc}  & 33.4 & 23.1 & 33.1 & 4.9  & 30.2 & 27.3 & 25.3 \\
Ours   & 36.5 & 21.1 & 36.2 & 12.7 & 31.9 & 30.2 & \textbf{28.1} \\
\midrule
\multicolumn{8}{c}{\textit{Occ3D-nus}~\cite{tian2023occ3d}~\textit{mIoU (\%) $\uparrow$}} \\
\midrule
FB-Occ~\cite{li2023fb} & 28.3 & 18.0 & 28.4 & 7.7 & 26.3 & 26.3 & 22.5 \\
COTR$^{\dagger}$~\cite{ma2024cotr}  & 30.3 & 19.2 & 30.4 & 8.8 & 27.7 & 28.7 & 24.2 \\  
STCOcc$^{\dagger}$~\cite{liao2025stcocc}& 33.9 & 20.1 & 31.8 & 11.0 & 28.7 & 29.3 & 25.8 \\
ALOcc~\cite{chen2025alocc}  & 36.9 & 22.2 & 35.3 & 10.1 & 29.9 & 31.6 & 27.7 \\
Ours   & 36.9 & 21.4 & 34.2 & 12.4 & 31.1 & 32.3 & \textbf{28.0} \\
\midrule
\multicolumn{8}{c}{\textit{Occ3D-nus}~\cite{tian2023occ3d}~\textit{RayIoU (\%) $\uparrow$}} \\
\midrule
FB-Occ~\cite{li2023fb} & 30.6 & 19.3 & 29.6 & 5.1  & 26.8 & 24.5 & 22.7 \\
OPUS-L~\cite{wang2024opus} & 37.8 & 24.4 & 34.4 & 5.7  & 22.6 & 28.7 & 25.6 \\
STCOcc~\cite{liao2025stcocc} & 33.5 & 19.4 & 31.3 & 7.8  & 31.0 & 26.4 & 24.9 \\
ALOcc~\cite{chen2025alocc}  & 36.0 & 22.5 & 34.1 & 6.5  & 30.2 & 26.9 & 26.0 \\
Ours   & 34.2 & 22.1 & 33.2 & 9.1   & 31.8 & 28.6 & \textbf{26.5} \\
\bottomrule
\end{tabular}}
\vskip-3ex
\end{table}

\noindent\textbf{Main Results on Occ3D-nus.}
As shown in Tab.~\ref{table:Occ3D_rayiou} and Tab.~\ref{table:Occ3D_miou}, our method achieves a RayIoU of $43.3$ and an mIoU of $44.6$ on Occ3D-nus~\cite{tian2023occ3d}.
Compared with the STCOcc~\cite{liao2025stcocc} baseline under the same ResNet-50 backbone and $256\times704$ input resolution, our method improves RayIoU by $+1.6$ and mIoU by $+2.4$.
Consistent gains are also observed for traffic-participant classes, including \emph{bicycle} ($+3.6$), \emph{car} ($+3.1$), \emph{truck} ($+2.6$), and \emph{pedestrian} ($+2.3$).
These improvements demonstrate competitive occupancy prediction for both overall scene structure and traffic participants, providing a reliable spatial basis for interaction reasoning.

\noindent\textbf{Main Results on nuScenes-C.}
We evaluate SelectOccFlow on nuScenes-C~\cite{xie2025benchmarking} under six visual corruptions, each with three severity levels. As shown in Tab.~\ref{tab:nuscenes_c_robustness}, our method achieves the best average performance under visual corruptions across both OpenOcc~\cite{tong2023scene} and Occ3D-nus~\cite{tian2023occ3d}. On OpenOcc, it improves the average OccScore from the previously reported $25.3$ to $28.1$. On Occ3D-nus, improved average mIoU and RayIoU further demonstrate better robustness to visual corruptions. The advantage becomes more evident under challenging corruptions such as \textit{Snow} and \textit{Motion}, where our method maintains higher prediction quality across the evaluated metrics. These results demonstrate improved robustness to visual corruptions, supporting reliable 3D perception under challenging driving conditions.

\noindent\textbf{Implications for Autonomous Driving.}
Across diverse visual conditions, especially challenging \textit{Snow} and \textit{Motion} cases, SelectOccFlow better preserves traffic-participant structure, while qualitative results show more faithful motion in ego-interaction regions. These gains are valuable for nearby pedestrians and other vulnerable road users, as reliable geometry and motion cues are critical for anticipating potential interactions. Overall, these results support more robust scene understanding and safer driving decisions.

\begin{figure*}[!t]
    \centering
    \includegraphics[width=0.93\textwidth]{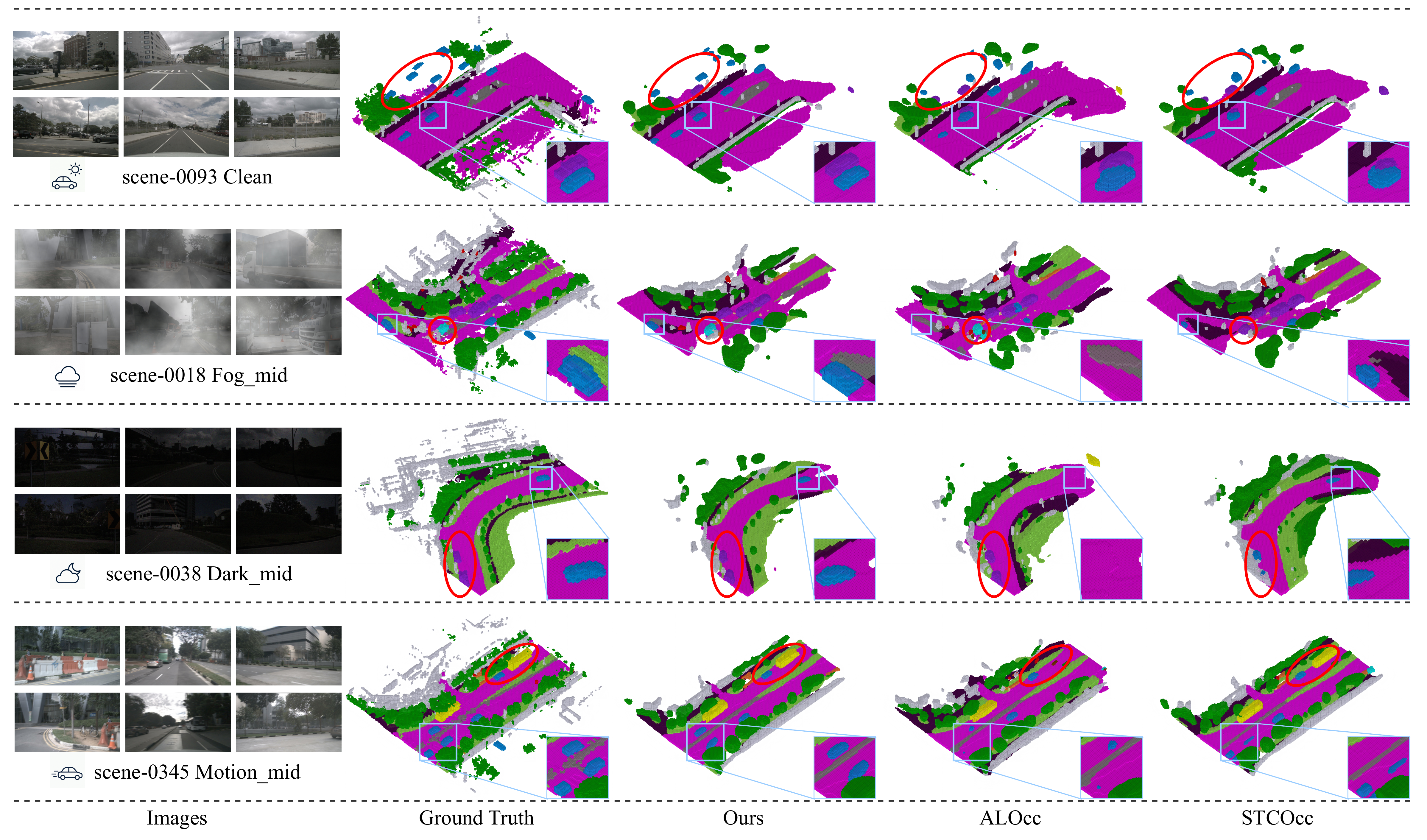}
    \vskip-2ex
    \caption{\textbf{Qualitative comparison of 3D semantic occupancy under clean and corrupted visual inputs.} \textit{Clean} samples are from OpenOcc~\cite{tong2023scene}, while \textit{Fog}, \textit{Dark}, and \textit{Motion} samples are from nuScenes-C~\cite{xie2025benchmarking}. From left to right, we show multi-view images, Ground Truth, Ours, ALOcc~\cite{chen2025alocc}, and STCOcc~\cite{liao2025stcocc}. Red circles and blue boxes highlight movable objects and enlarged regions, showing clearer boundaries and more complete structures.
    }
    \label{fig:fig3}
    \vskip-3ex
\end{figure*}
\subsection{Ablation Studies}
To analyze the contribution of each component, we conduct ablation experiments on the proposed SelectOccFlow framework using the OpenOcc benchmark~\cite{tong2023scene}.
\begin{table}[!t]
    \centering
    \caption{Ablation study of architectural components on OpenOcc~\cite{tong2023scene}.}
    \label{table:Ablation Study}
    \vskip-1ex
    \resizebox{\columnwidth}{!}{
    \begin{tabular}{c|ccc|ccc}
    \toprule
    & SGS & SCTA & ESA & RayIoU (\%) $\uparrow$ & mAVE $\downarrow$ & OccScore $\uparrow$ \\
    \midrule
    A&           &            &            & 38.9 & 0.525 & 39.8 \\
    \rowcolor{gray!8}B&\checkmark &            &            & 39.7 & 0.487 & 40.8 \\
    \rowcolor{gray!8}C&           & \checkmark &            & 41.5 & \underline{0.408} & 43.3 \\
    \rowcolor{gray!8}D&           &            & \checkmark & 39.4 & 0.512 & 40.3 \\
    \rowcolor{gray!15}E&           & \checkmark & \checkmark & \underline{42.7} & 0.419 & \underline{44.2} \\
    \rowcolor{gray!15}F&\checkmark &            & \checkmark & 40.2 & 0.489 & 41.3 \\
    \rowcolor{gray!15}G&\checkmark & \checkmark &            & 42.5 & 0.415 & 44.1 \\
    \rowcolor{gray!25}H&\checkmark & \checkmark & \checkmark & \textbf{43.1} & \textbf{0.394} & \textbf{44.9} \\
    \bottomrule
    \end{tabular}}
    \vskip-3ex
\end{table}

\noindent\textbf{Architectural Components.}
As shown in Tab.~\ref{table:Ablation Study}, configuration A serves as the baseline, achieving $38.9$ RayIoU, $0.525$ mAVE, and $39.8$ OccScore.
Comparing B, C, and D with A shows that all three components consistently improve performance, with SCTA providing the largest individual gain by increasing RayIoU to $41.5$ and reducing mAVE to $0.408$.
This indicates that temporal modeling benefits both spatial structure recovery and motion estimation.
The pairwise combinations further reveal clear complementarity among the components, with ESA $+$ SCTA and SGS $+$ SCTA achieving OccScores of $44.2$ and $44.1$, respectively, showing that spatial refinement and temporal aggregation can work synergistically.
With all three components enabled, configuration H achieves the best overall performance with $43.1$ RayIoU, $0.394$ mAVE, and $44.9$ OccScore.
These results demonstrate that SGS, SCTA, and ESA provide complementary refinements at different stages, jointly improving 3D occupancy and scene flow prediction.

\begin{figure*}[!t]
    \centering
    \includegraphics[width=0.9\textwidth]{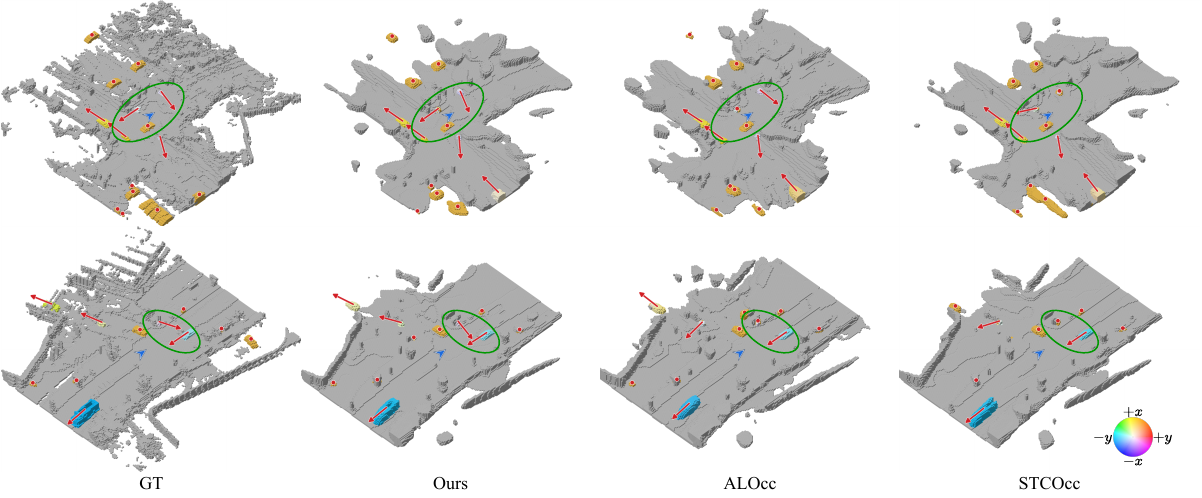}
    \vskip-2ex
    \caption{\textbf{Qualitative comparison of scene flow predictions on OpenOcc}~\cite{tong2023scene}\textbf{.} 
    Results from ALOcc~\cite{chen2025alocc} and STCOcc~\cite{liao2025stcocc} are included for comparison. Colors encode flow magnitude and direction, red arrows indicate motion direction, and colored dots denote movable objects moving slowly or remaining nearly stationary. Green ellipses highlight potential ego-vehicle interaction regions.
    }
    \label{fig:fig4}
    \vskip-3ex
\end{figure*}

\begin{table}[]
    \centering
    \caption{Ablation study of SCTA on OpenOcc~\cite{tong2023scene}.}
    \label{table:Ablation_sdtf}
    \vskip-1ex
    \resizebox{\columnwidth}{!}{
    \begin{tabular}{cc|ccc}
    \toprule
    Temporal & Motion & RayIoU (\%) $\uparrow$ & mAVE $\downarrow$ & OccScore $\uparrow$ \\ 
    \midrule
              &            & 40.2 & 0.489 & 41.3 \\
    \checkmark &            & \textbf{43.2} & \underline{0.432} & \underline{44.6} \\
    \rowcolor{gray!10}
    \checkmark & \checkmark & \underline{43.1} & \textbf{0.394} & \textbf{44.9} \\
    \bottomrule
    \end{tabular}
    }\vskip-3ex
\end{table}

\noindent\textbf{Ablation on SCTA.}
Tab.~\ref{table:Ablation_sdtf} analyzes the effects of temporal aggregation and the motion branch.
Adding temporal aggregation improves RayIoU from $40.2$ to $43.2$, reduces mAVE from $0.489$ to $0.432$, and raises OccScore from $41.3$ to $44.6$, showing that historical evidence benefits both occupancy and scene flow prediction.
Adding the motion branch further reduces mAVE from $0.432$ to $0.394$ while maintaining comparable RayIoU ($43.2$ \textit{vs.} $43.1$), increasing OccScore to $44.9$.
This indicates that the motion branch complements temporal aggregation by improving voxel-wise motion estimation while preserving occupancy quality.

\subsection{Qualitative Analysis}
We qualitatively evaluate 3D occupancy and scene flow prediction on the OpenOcc validation set~\cite{tong2023scene}.

\noindent\textbf{3D Occupancy Analysis.}
As shown in Fig.~\ref{fig:fig3}, red circles and enlarged blue boxes highlight representative movable objects under clean and corrupted inputs.
Under \textit{Clean} conditions, our method recovers more complete object structures with clearer boundaries.
Under \textit{Fog}, \textit{Dark}, and \textit{Motion} corruptions, the highlighted foreground objects remain better preserved, while competing methods tend to exhibit missing regions or less distinct boundaries.
These results demonstrate more robust foreground geometry recovery under visual corruptions, supporting reliable 3D perception of traffic participants and scene structure in challenging driving scenes.

\noindent\textbf{Scene Flow Analysis.}
Building on the occupancy predictions, Fig.~\ref{fig:fig4} further visualizes scene flow, with green ellipses highlighting regions where nearby movable objects may interact with the ego vehicle.
Within these regions, our predictions more closely match the ground truth in flow direction and magnitude than ALOcc~\cite{chen2025alocc} and STCOcc~\cite{liao2025stcocc}, while slow-moving and nearly stationary objects are also characterized more faithfully.
Accurate scene flow prediction in such interaction regions provides complementary dynamic cues beyond occupancy, supporting a more reliable understanding of traffic participants and their potential interactions with the ego vehicle in dynamic driving scenes.
        \section{Conclusion}

We presented SelectOccFlow, a selective spatiotemporal aggregation framework for 3D semantic occupancy and scene flow prediction. The proposed framework progressively refines image, temporal, and voxel evidence according to semantic compatibility, temporal reliability, and spatial structure, leading to improved dynamic 3D scene representation.
Future work will explore deployment-oriented extensions of the framework and tighter integration with downstream interaction reasoning and planning for autonomous driving.

        \bibliographystyle{IEEEtran}
        \bibliography{bib}
\end{document}